\pdfoutput=1

\documentclass[11pt]{article}

\usepackage[]{naacl2021}

\usepackage{times}
\usepackage{latexsym}

\usepackage[T1]{fontenc}

\usepackage[utf8]{inputenc}

\usepackage{microtype}

\title{Never guess what I heard... Rumor Detection in Finnish News: a Dataset and a Baseline}

\author{Mika Hämäläinen, Khalid Alnajjar, Niko Partanen and Jack Rueter \\
  Department of Digital Humanities\\
  University of Helsinki\\
  \texttt{firstname.lastname@helsinki.fi} }

\begin{document}
\maketitle
\begin{abstract}
This study presents a new dataset on rumor detection in Finnish language news headlines. 
We have evaluated two different LSTM based models and two different BERT models, and have found very significant differences in the results. A fine-tuned FinBERT reaches the best overall accuracy of 94.3\% and rumor label accuracy of 96.0\% of the time. However, a model fine-tuned on Multilingual BERT reaches the best factual label accuracy of 97.2\%. Our results suggest that the performance difference is due to a difference in the original training data. Furthermore, we find that a regular LSTM model works better than one trained with a pretrained word2vec model. These findings suggest that more work needs to be done for pretrained models in Finnish language as they have been trained on small and biased corpora.
\end{abstract}

\section{Introduction}

Contemporary online news media contains information from more and less reputable sources, and in many cases the reliability of individual news articles can be very questionable. 
This has far reaching impact on society and can even influence decision making, as everyone continuously encounters such material online. This is a real issue as identified by \citet{paskin2018real}. In their study, they found out that participating people could only, on the average, distinguish fake news from real news half of the time, and none of the participants was able to identify all samples of fake and real news correctly.

Ever since the 2016 US elections fake news and misinformation have become a hot topic in the English speaking world \cite{10.1257/jep.31.2.211}, however other countries are no less immune to the spread of such information. In this study we present for the first time a method for rumor detection in Finnish news headlines. Finnish language has not yet received any research interest in this topic, and for this reason we also propose a new dataset\footnote{Our dataset is freely available for download on Zenodo https://zenodo.org/record/4697529} for the task. We treat this task as a classification problem where each news headline is categorized as being either rumorous or factual.

Rumor detection is a very challenging task, and we believe that truly satisfactory results need to leverage other methods than only natural language processing. 
Whether a given text is a rumor or not is very strongly connected to real world knowledge and continuously changing world events that we don't believe that this can be solved within the analysis of individual strings without a larger context. However, if we can perform even a rough classification at this relatively simple level, this could be used as one step in more robust and complex implementations. Therefore, our initial approach should be seen as a baseline for future implementations, while it can be seen as an important advancement in this work, and in this case it is the starting point, as related work in matters of this topic for Finnish remains nonexistent. 

\section{Related Work}

Rumor detection has in recent years become an active topic of investigation, especially due to the complex influence it has on modern societies through social media. There has been other work on rumor detection for languages other than English as well. 
\citet{alzaninEtAl2019a} studied rumor detection in Arabic tweets and \citet{chernyaevEtAl2020a} in Russian tweets. 
Recently, \citet{keEtAl2020a} has also presented a method for rumor detection in Cantonese. A closely related topic, stance detection, has been studied in a comparable corpus of French Tweets \cite{evrardEtAl2020a}. In this section, we describe some of the related work in more detail.

\citet{rubin-etal-2016-fake} harnessed satire in the task of fake news detection, in their study, this figure of language, that has also sparked research interest in detection \cite{li2020multi} and generation \cite{alnajjar2018master} on its own, was useful in detecting fake news. They proposed an SVM (support vector machines) approach capturing five features: \textit{Absurdity}, \textit{Humor}, \textit{Grammar}, \textit{Negative} \textit{Affect} and \textit{Punctuation}. The idea of satire in fake news detection was also studied later on by \citet{levi-etal-2019-identifying}.

Tree LSTMs have been used recently in rumor detection \cite{kumar-carley-2019-tree}. They train the models on social media text which contains interactions as people reply to statements either providing supporting or contradicting statements. Their model is capable of taking these replies into account when doing predictions.

\citet{sujana-etal-2020-rumor} detect rumors by using multiloss hierarchical BiLSTM models. The authors claim the hierarchical structure makes it possible to extract deep information form text. Their results show that their model outperforms a regular BiLSTM model.

Previous work on Finnish news materials include a study by \cite{ruokolainenEtAl2019a}, where the articles were annotated for named entities. 
In addition, other researchers have targeted Finnish news materials, especially historical newspapers that are openly available. 
Furthermore, \cite{melaEtAl2019a} has studied NER (named entity recognition) in the context of historical Finnish newspapers, and \cite{marjanenEtAl2020a,hengchen2019data} have tested methods for analyzing broader changes in a historical newspaper corpus. Our work departs from this, as we focus on modern newspaper headlines.

Additionally, to our knowledge there has not been any previous work on rumor detection for Finnish, which makes our work particularly novel and needed. 

\section{Data}

We collect data from a Finnish news aggregation website\footnote{https://www.ampparit.com/}, in particular, we crawl the news headlines in the rumor category to form samples of rumor data. In addition, we crawl the headlines in the category news from Finland to compile a list of headlines that do not contain rumors but actual fact-based news stories. This way we have gathered 2385 factual and 1511 rumorous headlines totaling to 3896 samples. As a preprocessing step, we tokenize the headlines with NLTK \cite{10.3115/1118108.1118117} word tokenizer.

\begin{table}[]
\centering
\begin{tabular}{|l|l|l|}
\hline
           & rumor & factual \\ \hline
train      & 1057  & 1669    \\ \hline
test       & 227   & 358     \\ \hline
validation & 227   & 358     \\ \hline
\end{tabular}
\caption{The size of the data splits on a headline level}
\label{tab:splits}
\end{table}

We shuffle the data and split it 70\% for training, 15\% for validation and 15\% for testing. The actual sizes can be seen in Table \ref{tab:splits}. We use the same splits for all the models we train in this paper. An example of the data can be seen in Table \ref{tab:examples}. 
The dataset has been published with an open license on Zenodo with a permanent DOI\footnote{https://zenodo.org/record/4697529}. The splits used in this paper can be found in the dataset\_splits.zip file. 

\begin{table*}[]
\centering
\begin{tabular}{|l|l|}
\hline
Headline                                                                                                                                                                                       & Rumor \\ \hline
\begin{tabular}[c]{@{}l@{}}Tutkimus: Silmälaseja käyttävillä ehkä pienempi riski koronatartuntaan\\ \textit{Study: People wearing eyeglasses may have a smaller risk of getting corona}\end{tabular}        & true  \\ \hline
\begin{tabular}[c]{@{}l@{}}Koronaviruksella yllättävä sivuoire - aiheutti tuntikausien erektion\\ \textit{Coronavirus has a surprising symptom - caused an erection that lasted for hours}\end{tabular} & true  \\ \hline
\begin{tabular}[c]{@{}l@{}}Korona romahdutti alkoholin matkustajatuonnin\\ \textit{Corona caused a collapse in traveler import of alcohol}\end{tabular}                                                & false \\ \hline
\begin{tabular}[c]{@{}l@{}}Bussimatka aiheutti 64 koronatartuntaa\\ \textit{A bus trip caused 64 corona cases}\end{tabular}                                                                            & false \\ \hline
\end{tabular}
\caption{Examples of rumorous and factual headlines related to COVID-19 from the corpus}
\label{tab:examples}
\end{table*}

\section{Detecting Rumors}

In this section, we describe the different methods we tried out for rumor detection. We compare LSTM based models with transfer learning on two different BERT models.

We train our first model using a bi-directional long short-term memory (LSTM) based model \cite{hochreiter1997long} using OpenNMT \cite{opennmt} with the default settings except for the encoder where we use a BRNN (bi-directional recurrent neural network) \cite{schuster1997bidirectional} instead of the default RNN (recurrent neural network). We use the default of two layers for both the encoder and the decoder and the default attention model, which is the general global attention presented by \citet{luong2015effective}. The model is trained for the default 100,000 steps. The model is trained with tokenized headlines as its source and the rumor/factual label as its target.

We train an additional LSTM model with the same configuration and same random seed (3435) with the only difference being that we use pretrained word2vec embeddings for the encoder. We use the Finnish embeddings provided by \cite{kutuzov2017word}\footnote{http://vectors.nlpl.eu/repository/20/42.zip}. The vector size is 100 and the model has been trained with a window size 10 using skipgrams on the Finnish CoNLL17 corpus.

In addition, we train two different BERT based sequence classification models based on the Finnish BERT model FinBERT \cite{virtanen2019multilingual} and Multilingual BERT \cite{devlin2019bert}, which has been trained on multiple languages, Finnish being one of them. We use the transformers package \cite{wolf-etal-2020-transformers} to conduct the fine tuning. As hyperparameters for the fine tuning, we use 3 epochs with 500 warm-up steps for the learning rate scheduler and 0.01 as the strength of the weight decay.

This setup takes into account the current state of the art at the field, and uses recently created resources such as Finnish BERT model, with our own custom made dataset. 
Everything is set up in an easily replicable manner, which ensures that our experiments and results can be used in further work on this important topic. 

\section{Results}

 
In this section, we present the results of the models, in addition, we explain why these result were obtained by contrasting the task into the training data of each pretrained model. The accuracies of the models can be seen in Table~\ref{tab:results}. 

\begin{table}[]
\centering
\begin{tabular}{|l|l|l|l|}
\hline
                  & Overall         & Factual         & Rumor           \\ \hline
LSTM              & 84.9\%          & 93.2\%          & 71.8\%          \\ \hline
LSTM + word2vec   & 71.6\%          & 94.4\%          & 35.6\%          \\ \hline
FinBERT           & \textbf{94.3\%} & 93.2\%          & \textbf{96.0\%} \\ \hline
Multilingual BERT & 91.8\%          & \textbf{97.2\%} & 83.3\%          \\ \hline
\end{tabular}
\caption{Overall and label level accuracies for each model }
\label{tab:results}
\end{table}

The results vary greatly, with tens of percentages between different approaches. 
It is important to note that while FinBERT gets the best overall accuracy and the best accuracy in predicting rumorous headlines correctly, it does not get the best accuracy in predicting factual headlines correctly, as it is actually Multilingual BERT that gets the best accuracy for factual headlines. This makes us wonder why this might be so. When we look at the training data for these models, we can see that Multilingual BERT was trained on Wikipedia\footnote{https://github.com/google-research/bert/blob/master/multilingual.md}, whereas FinBERT was mainly trained on data from an internet forum, Suomi24\footnote{https://keskustelu.suomi24.fi/}, that is notorious for misinformation, (33\% of the data) and Common Crawl\footnote{https://commoncrawl.org/} (60\% of the data). Only 7\% of the training data comes from a news corpus. When we put the results into perspective with the training data, it is not at all the case, as the authors of FinBERT claim in their paper: \textit{"The results indicate that the multilingual models fail to deliver on the promises of deep transfer learning for lower-resourced languages"} \cite{virtanen2019multilingual}. Instead, based on our results, it is only evident that Multilingual BERT outperforms FinBERT on factual headlines as its training data was based on an encyclopedia, and that FinBERT is better at detecting rumors as its training data had a large proportion of potentially rumorous text from Suomi24 forum.

In the same fashion, we can explain the results of the LSTM models. A great many papers \cite{qi-etal-2018-pre,panchendrarajan-amaresan-2018-bidirectional,khalid-embeddings} have found that pretrained embeddings improve prediction results when used with an LSTM model, however, in our case, we were better off without them. While the data description \cite{zeman-EtAl:2017:K17-3} was not clear on what the data of the pretrained word2vec model consists of (apart from it being from Common Crawl), we can still inspect the overlap in the vocabulary of the training data and the pretrained model. Our training and validation datasets contain 17,729 unique tokens, out of which 5,937 were not present in the pretrained model. This means that approximately 33\% of the tokens in our dataset were simply not present in the pretrained model. 

This is partially due to the English driven tradition of not lemmatizing pretrained models, however, for a language such as Finnish this means that a simple overlap in vocabulary is not enough, instead one would even need to have overlap in the syntactic positions where the words have appeared in the data of a pretrained model and in the training data of the model that would utilize the pretrained embeddings. It is important to note that the pretrained word2vec model does not have a small vocabulary either (2,433,286 tokens).

In order to study whether the issue arises from the fact that the word2vec model is not lemmatized or from the fact that its training data was from a different domain, we conduct a small experiment. We lemmatize the words in our training and validation dataset and the words in the vocabulary of the word2vec model by using the Finnish morphological FST (finite-state transducer) Omorfi \cite{pirinen2015development} through UralicNLP\footnote{We use the dictionary forms model} \cite{uralicnlp_2019}. After lemmatization, our corpus contains 10,807 unique lemmas, 2,342 out of which are still not in the lemmatized vocabulary of the word2vec model. This means that even on the lemma level, 21.7\% of the words are not covered by the word2vec model. The lemmatized vocabulary size of the pretrained model is 576,535 lemmas. It is clear that a model leveraging from sub-word units could not alleviate the situation either, as such models are mainly useful to cope with inflectional forms, but not with completely new words that merely look familiar on the surface.

\section{Conclusions}

Our study shows that with the tested settings it is possible to differentiate the rumor and non-rumor categories with a very high accuracy. As the experiment setup was relatively simple, yet elegant, we believe that similar results can also be repeated for other languages for which rumor detection systems have not yet been created. The experiments reported here are just one part in creating such a system for Finnish language. We believe that the path towards more thorough solutions lies in larger manually annotated datasets that contain even more variation than the materials we have now used. Although, some of these datasets could be automatically generated by using Finnish semantic databases \cite{hamalainen2018extracting} and syntax realization \cite{hamalainen2018development} in conjunction with existing Finnish news headline generation methods \cite{alnajjar2019no}.

Possibly the most relevant finding of our study lies, however, in the results we detected with different BERT models and were able to connect into the differences in training data. These findings are important much beyond just rumor detection, which is only one domain where these models are being continuously used. As the question of training data seemed to be an important one also for the word2vec model in the LSTM experiment, we can only conclude that the level of the existing pre-trained models for Finnish is not good enough for them to work in many different domains. This is not a question of Finnish being "low resourced" (see \citealt{mika-endangered}), as huge amounts of text exist in Finnish online, but more of a question of not enough academic interest in producing high-quality models. This is something we will look into in the future.

Further work is needed from a qualitative perspective to see what exactly leads to a certain classification, and which kind of error types can be detected. Since the classification was done solely based on linguistic features of the text, represented by the strings, we must assume there are lexical and stylistic differences that are very systematic. Not unlike in the case of the existing methods for rumor detection, our models did not have access to any real world knowledge about the rumors or factual and non-factual information at the time when the headlines were written. It is obvious that a very well functioning system can only be built in connection to this kind of sources, as the fact that something is a rumor is ultimately connected to the content and real world knowledge, and not just the words in the string.  However, we argue that our system could already be useful in a rough classificatory tasks where rumor containing news could be automatically selected for manual verification, or for verification with a more specialized neural network. Naturally further work also has to take into account more non-rumor text types and genres, so that certain degree of robustness can be reached. 

\bibliography{anthology,custom}
\bibliographystyle{acl_natbib}

\end{document}